\documentclass[sn-mathphys,Numbered]{sn-jnl}

\usepackage{graphicx}%
\usepackage{multirow}%
\usepackage{amsmath,amssymb,amsfonts}%
\usepackage{amsthm}%
\usepackage{mathrsfs}%
\usepackage[title]{appendix}%
\usepackage{xcolor}%
\usepackage{textcomp}%
\usepackage{manyfoot}%
\usepackage{booktabs}%
\usepackage{algorithm}%
\usepackage{algorithmicx}%
\usepackage{algpseudocode}%
\usepackage{listings}%

\theoremstyle{thmstyleone}%
\theoremstyle{thmstyletwo}%

\theoremstyle{thmstylethree}%

\begin{document}

\title[Article Title]{\vspace{-1cm}Camera Calibration for the Surround-View System: A Benchmark and Dataset}


\author[]{\fnm{Leidong} \sur{Qin}}\email{21120295@bjtu.edu.cn}

\author*[]{\fnm{Chunyu} \sur{Lin}*}\email{cylin@bjtu.edu.cn}

\author[]{\fnm{Shujun} \sur{Huang}}\email{shujuanhuang@bjtu.edu.cn}

\author[]{\fnm{Shangrong} \sur{Yang}}\email{sr\_yang@bjtu.edu.cn}

\author[]{\fnm{Yao} \sur{Zhao}}\email{yzhao@bjtu.edu.cn}

\affil{\orgdiv{Institute of Information Science}, \orgname{Beijing Jiaotong University}, \orgaddress{\street{Beijing Key Laboratory of Advanced Information Science and Network}, \city{Beijing}, \postcode{100044}, \country{China}}}

\abstract{Surround-view system (SVS) is widely used in the Advanced Driver Assistance System (ADAS). SVS uses four fisheye lenses to monitor real-time scenes around the vehicle. However, accurate intrinsic and extrinsic parameter estimation is required for the proper functioning of the system. At present, the intrinsic calibration can be pipeline by utilizing checkerboard algorithm, while extrinsic calibration is still immature. Therefore, we proposed a specific calibration pipeline to estimate extrinsic parameters robustly. This scheme takes a driving sequence of four cameras as input. It firstly utilizes lane line to roughly estimate each camera pose. Considering the environmental condition differences in each camera, we separately select strategies from two methods to accurately estimate the extrinsic parameters. To achieve accurate estimates for both front and rear camera, we proposed a method that mutually iterating line detection and pose estimation. As for bilateral camera, we iteratively adjust the camera pose and position by minimizing texture and edge error between ground projections of adjacent cameras. After estimating the extrinsic parameters, the surround-view image can be synthesized by homography-based transformation. The proposed pipeline can robustly estimate the four SVS camera extrinsic parameters in real driving environments. In addition, to evaluate the proposed scheme, we build a surround-view fisheye dataset, which contains 40 videos with 32,000 frames, acquired from different real traffic scenarios. All the frames in each video are manually labeled with lane annotation, with its GT extrinsic parameters. Moreover, this surround-view dataset could be used by other researchers to evaluate their performance. The dataset will be available soon.}




\maketitle

\section{Introduction}\label{sec1}

Surround-view is increasingly popular in advanced driver assistance system (ADAS) \cite{bib1}\cite{bib2} A four-camera SVS system is shown in Fig.1. The Surround-view generates a  360-degree image around the vehicle, providing the driver with a comprehensive view of the environment without any blind spots. In addition, the Surround-view system is widely used in various automatic driving computer vision tasks, including traffic sign recognition parking space detection \cite{bib4}\cite{bib5}\cite{bib6}. Typically, the onboard cameras are installed at the front, rear, left, and right sides of the vehicle,  near the license plate. The images captured by onboard cameras are then used to generate auxiliary views of the vehicle environment.

Achieving seamless stitching of all captured images requires accurate calibration of the multi-camera system. Inaccurate calibration parameters can lead to a false perception of surroundings, which is dangerous for vehicle control. While intrinsic calibration techniques are mature, extrinsic parameters are varied due to the tiny camera motion. As the vehicle moves, the camera can slowly accumulate extrinsic changes from vibrations such as engine vibration, door opening and closing, extreme wind, and road bumps. Therefore, the pose and position of the onboard camera need to be re-estimated to ensure stitching performance. In most commercial solutions, drivers have to rely on professional factories or workers for extrinsic calibration, which can be time-consuming and labor-intensive.  Manufacturers have a demand for effective extrinsic calibration without human intervention. In recent years, there are some self-calibration schemes for various realistic scenarios \cite{bib24}\cite{bib25}\cite{bib26}\cite{bib27}. However, the dataset of traffic scenarios with surround-view is still insufficient. To address this research gap, we collect a new dataset that consists of traffic scenes with lane lines and proposed a self-calibration pipeline. In summary, our contributions mainly consist of three aspects:

1)	An extrinsic calibration scheme is proposed, which utilizes lane line and ground texture. First, our method detects and filters lane lines nearby the vehicle. Based on geometric constraints, we makes a rough estimation for camera pose by utilizing the direction of lane lines. Second, we align the ground texture of adjacent cameras to achieve accurate extrinsic estimation. 

2)	Our method corrects pose through mutual iterating of lane line projection constraints and lane re-detection. Rough lane detection results in inaccurate pose estimation.  By projecting the frame to the rough ground plane with the rough pose, lane marking can be relocated more accurately. As a result, the pose can be iterated with updated lane marking. Our method is robust and can maintain subpixel accuracy for long-range lane marking detection. 

3)	We propose a surround-view video dataset with ground truth (GT). The dataset is collected from various traffic scenarios with lane lines in different environmental conditions. It contains 40 sets of videos collected by fisheye cameras. All video frames are manually annotated with high-quality lane points and camera extrinsic parameters.

The proposed method requires following assumptions as prerequisites:

    1)	Intrinsic parameters and the position of the camera are already known.
    
    2)	The road surface is flat and straight, besides the driving direction is always parallel to the lane line.The angle between the front direction of the vehicle and the lane lines is supposed to be from -2 to 2 degree. In our calibration pipeline, the range of roll and yaw angle between front direction of the vehicle and the lane lines were supposed to be from -10 to 10 degrees. The range of pitch angle is from 20 to 90 degrees. For each camera, The Intrinsic parameters (including distortion coefficients) are required, and the error of the x-coordinate should be within -0.2 and +0.2 meters. The error of the z-coordinate (the height of camera) should be within -0.2 and +0.2 meters.
    
SVS system is usually consist of four or six fisheye cameras. Our paper is aimed at the four-camera system. The proposed method is fully automatic. In traffic scenarios, it can maintain stable estimation under various environmental conditions.

\begin{figure}[h]%
\centering
\includegraphics[width=1.0\textwidth]{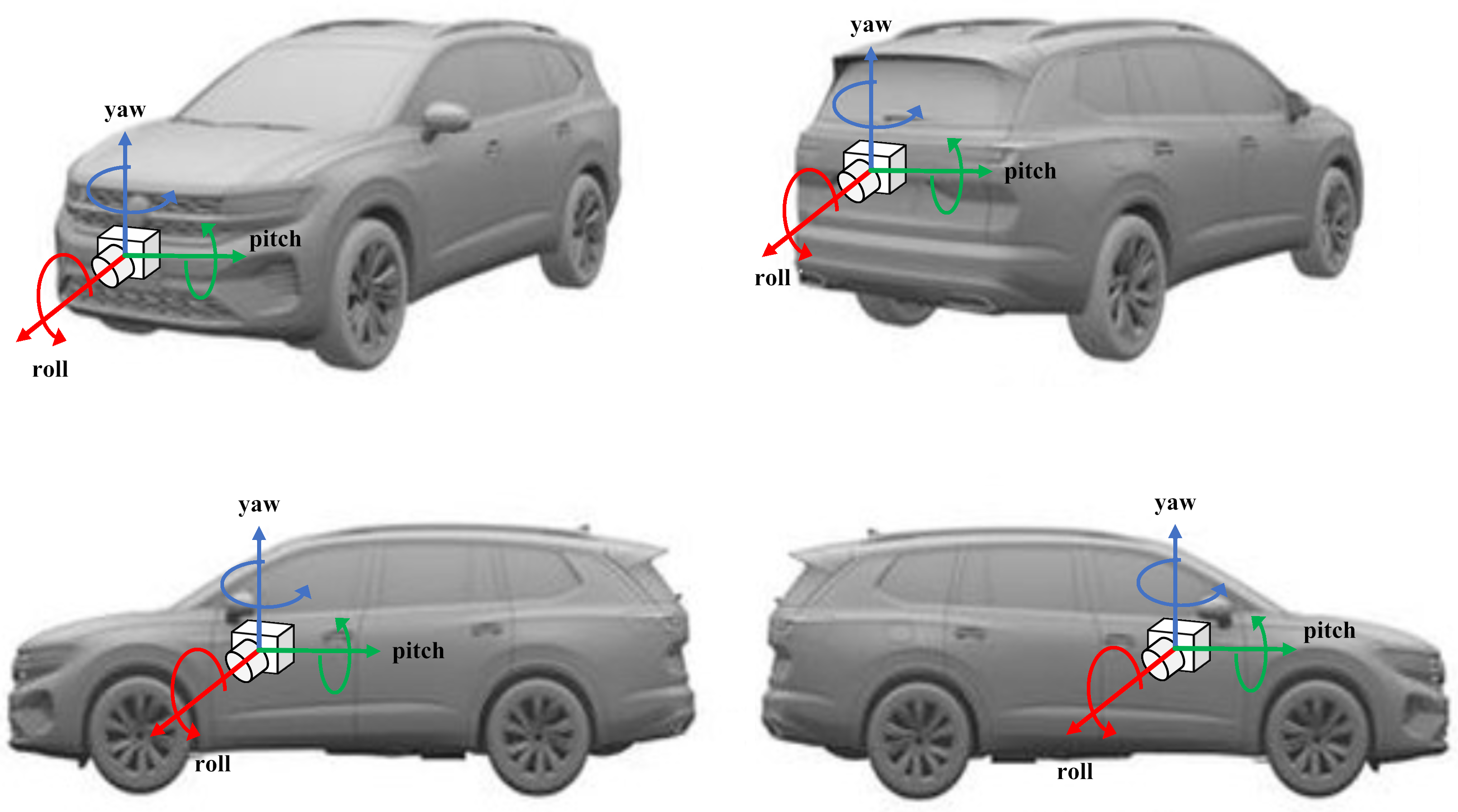}
\caption{ \centering A four-camera SVS system.}\label{fig1}
\end{figure}

\section{Related Works}\label{sec2}

\subsection{Surround-view datasets}\label{subsec2}

The Woodscape dataset\cite{bib31} is the first fisheye dataset that includes about 10,000 images with instance-level semantic annotation. It was collected from the street road and parking lot. Notably, the images from the dataset are not continuous frame by frame and are selected periodically. The entire dataset were collected by the same vehicle, resulting in only one set of extrinsic parameters.

The Tongji Surround-view dataset \cite{bib3} is a large-scale campus  scene collected by an electric car equipped with four cameras. It mainly composed of two parts: calibration site images and natural scenes.  The original dataset contains 19,078 groups of fisheye images with only one set of extrinsic parameters.  

\subsection{Lane detection}\label{subsec2}

Lane markings are general elements in traffic scenes. Utilizing the geometric constraints, pose can be estimated by a pair of lane marking in frames.

Early lane detection methods were primarily based on hand-crafted features, such as color \cite{bib7}\cite{bib8}, edge \cite{bib9}\cite{bib10}, and texture \cite{bib11}. Recently, the methods of deep learning \cite{bib12}\cite{bib13} have significantly improved lane detection performance. In VPGNet \cite{bib14}, vanishing points were utilized to conduct multi-task network training for lane detection. SCNN \cite{bib15} specifically considers the thin-and-long shape of lanes by passing messages between adjacent rows and columns at a feature layer. SAD \cite{bib16} and inter-region affinity KD \cite{bib17} further adopt the knowledge distillation to improve lane detection. PolyLaneNet \cite{bib18} formulates the instance-level lane detection as a polynomial regression problem, and UFSA \cite{bib19} provides ultra-fast lane detection by dividing the image into grids and scanning grids for lane locations.  

Due to the limitation of computation performance, on-board processors cannot apply to deep leaning tasks. To achieve a more accurately extrinsic calibration, we propose a method that provide subpixel lane detection in the middle distance.

\subsection{Pattern-based calibration methods}\label{subsec2}

A pattern-based approach estimates camera parameters using special patterns, including corners, circles, or lines. Since a pattern-based approach uses precisely drawn patterns with known configurations, it is possible to accurately estimate the camera parameters, making  it suitable for accurate calibration for a surround-view camera system. Some pattern-based calibration methods place calibration patterns in the overlapped fields of the cameras \cite{bib20}\cite{bib21}. Methods in \cite{bib22} use factorization-based methods by placing calibration patterns between adjacent groups of cameras. However, these calibration method requires a checkerboard, making them suitable for the factory but not for a private use, as they involve lots of human interventions.

\subsection{Self-calibration methods}\label{subsec2}

In \cite{bib23}, Zhao et al. first detected multiple vanishing points of lane markings on the road via the weighted least squares method. With the estimated vanishing points, the pose of the multi-camera system relative to the world coordinate system was solved. Choi et al. \cite{bib24} also designed a lane-line based extrinsic self-calibration pipeline for the surround-view case, in which the SVS was calibrated by aligning lane markings across images of adjacent cameras. However, this method relies on high accuracy of lane marking detection and world coordinate of cameras. 

There are some self-calibration schemes that are applicable to the SVS, including Liu et al.’s method \cite{bib25} and Zhang et al.’s \cite{bib26}\cite{bib27}. They deeply dissected the online extrinsic correction problem and offered effective solutions. Liu et al. \cite{bib25} proposed two models, the “Ground Model” and the “Ground-Camera Model”, which correct extrinsic by minimizing photometric errors with the steepest descent \cite{bib25}. Zhang et al.\cite{bib26} designed a novel model, the bi-camera model, to construct the least-square errors \cite{bib28} on the imaging planes of two adjacent cameras and then optimize camera poses by the LM (Levenberg-Marquardt) algorithm \cite{bib29}. And they further improved their work in \cite{bib27} by utilizing multiple frames selected in a local window rather than a single frame to build the overall error, thus improving the system’s robustness. However, the above three studies \cite{bib25}\cite{bib26}\cite{bib27} focused on online correction rather than calibration, and required a rough initial extrinsic as the input, which  limited the application.

\section{Proposed Method}\label{sec3}

Figure. 2 shows the flowchart of the proposed method. First, the method detects lane markings from images captured by four cameras. After locating the lane markings, the method calibrates the extrinsic parameter the of the front and rear cameras. Then it processes the left and right cameras using the estimated  of the front and rear cameras. In the following  section, further details of each step will be introduced. 

\begin{figure}[h]%
\centering
\includegraphics[width=1.0\textwidth]{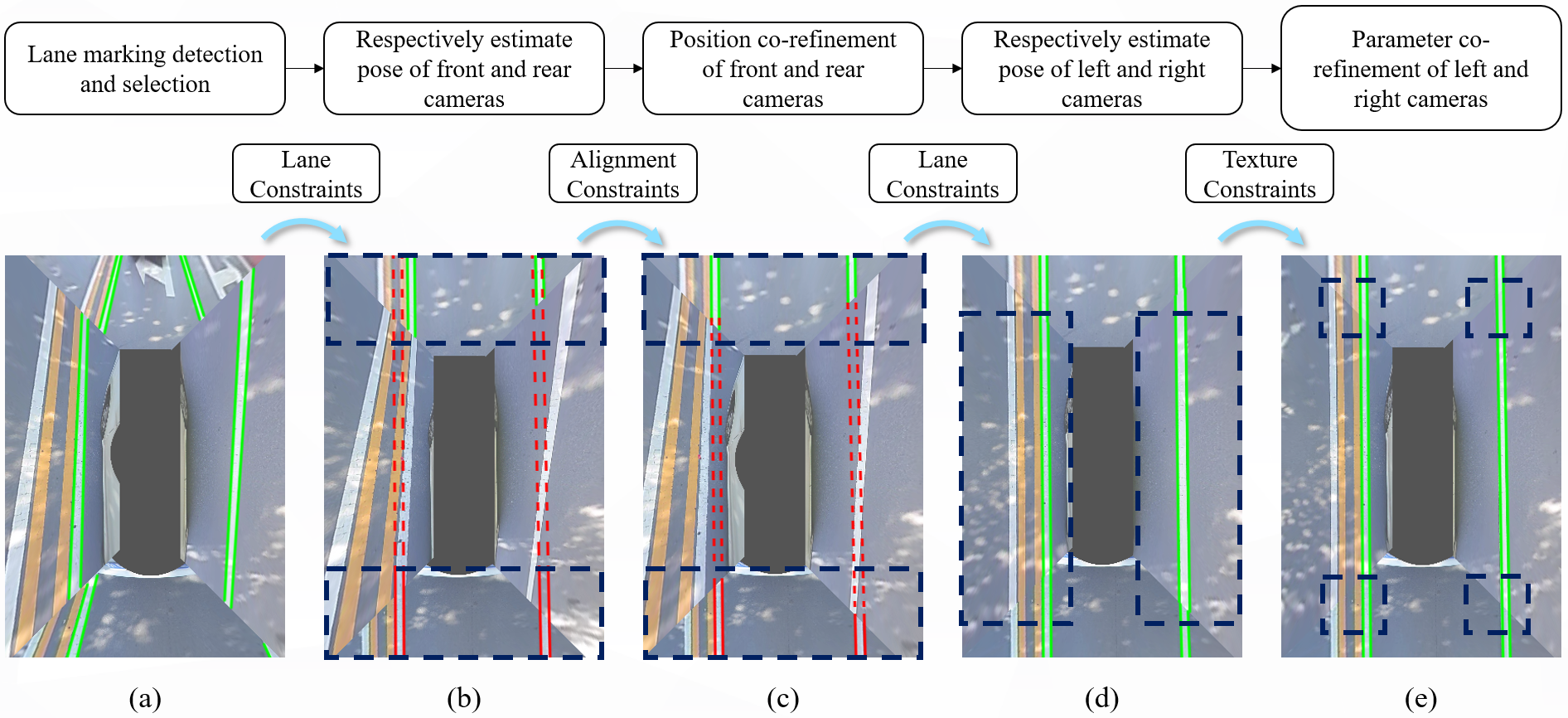}
\caption{Flowchart of the proposed method. (a) to (e) respectively show the surround-view generated by extrinsic parameters of each step.}\label{fig2}
\end{figure}

\subsection{Lane detection and selection}\label{subsec3}

We utilize Kannala-Brandt’s distortion model to undistort fisheye images and Canny’s edge to detect lines. The proposed method employs Hough line detection on the undistorted image to detect lane marking. Since the detection results may include falsely detected lane markings, a series of measures are taken to filter the required lane markings. First, we used a vanishing point (VP) to reject the outer lines. The VP can be calculated by \cite{bib30}. The Lane marking whose perpendicular distance to the vanishing point is less than a specified threshold are preserved. The overall aim is to reject the lines that are far away from the vanishing point.  Considering the left and right cameras cannot capture the lane markings below the vehicle, a RoI (region of interest) is set to reject such markings. Since the front and rear cameras required only one pair of lane marking, the pair of lane marking nearest to the camera is selected for estimation. As for the left and right cameras, it only requires one lane marking.

\begin{figure}[h]%
\centering
\includegraphics[width=0.9\textwidth]{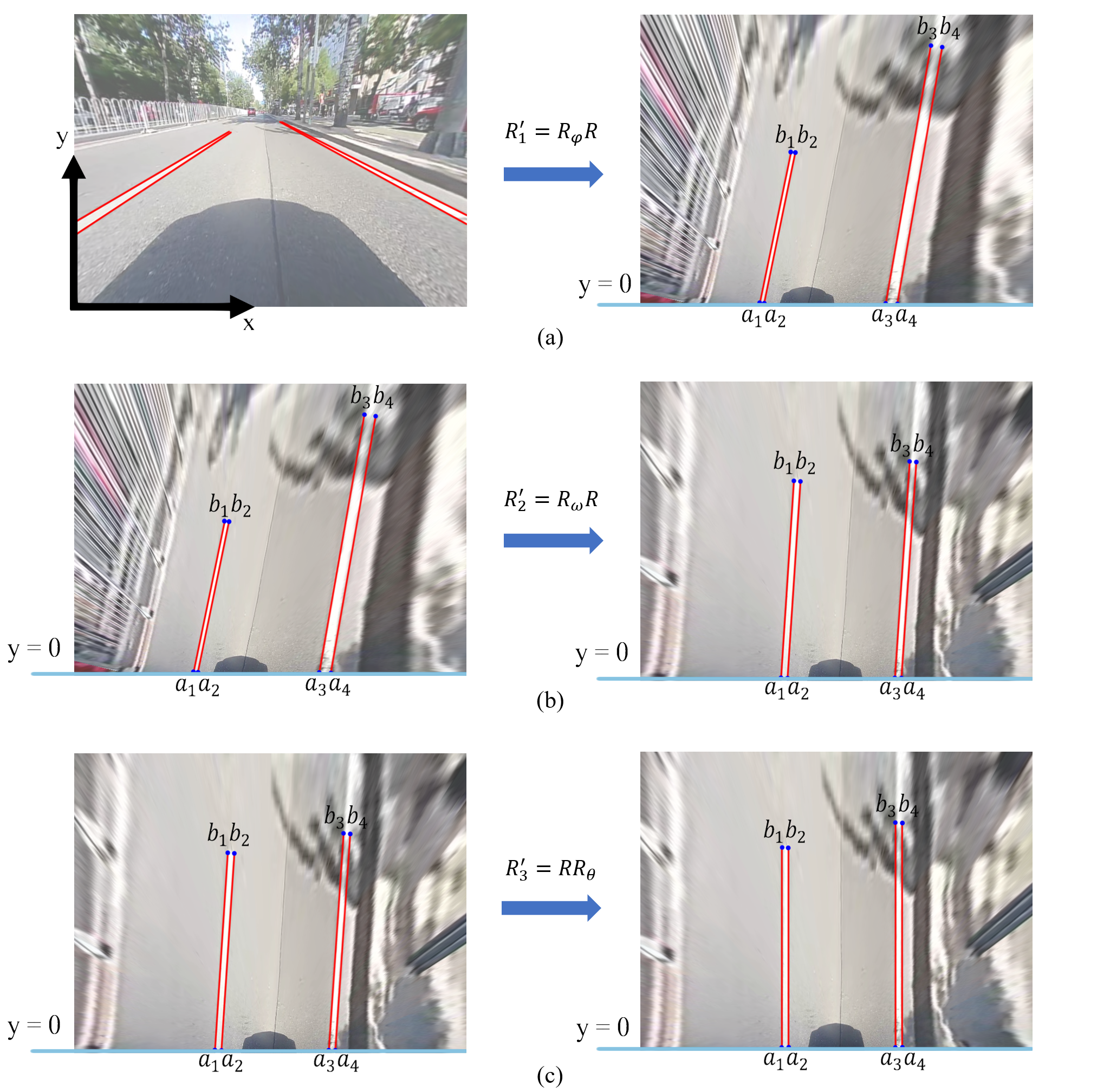}
\caption{Algorithm of pose estimation for front and rear camera.  (a). Sub-step of approaching pitch angle. (b) Sub-step of approaching yaw angle . (c) Sub-step of approaching roll angle.}\label{fig3}
\end{figure}

\subsection{Front and rear cameras Calibration}\label{subsec3}

The proposed method uses geometric constraints of lane lines in the projection plane to iteratively estimate camera pose. 

We propose an algorithm to iteratively estimate camera pose. The algorithm consisted of three sub-step and each sub-step separately estimate pitch, yaw and roll angles. Assuming $ P' $ is the point in the undistorted image and the corresponding point in the surround-view image is $ P $, $ P $ can be transformed by the homography $ H $ as follows,

\begin{equation}
   P^{'}  = H^{-T}P = \left ( K_2RK_1^{-1} \right )^{-T}P 
 ,\label{eq1}
\end{equation}

where $ K_1 $ is a 3x3 matrix representing the real camera’s intrinsic parameters  and $ K_2 $ is a 3x3 matrix representing the intrinsic parameters of the virtual camera on the surround-view plane. $ R $ is the rotation matrix. $ I^{’} $ is the point on the surround-view plane.

We design three kinds of cost function to estimate pitch, roll and yaw angle separately. Figure 3 illustrates their functionality. Our algorithm first minimizes the cost $ c_{ \varphi } $ by updating rotation matrix iteratively with,

\begin{equation}
   R_{1}^{'} =R_{\varphi } R=\begin{bmatrix}
     1 & 0 & 0\\
     0 & \cos\varphi   & \sin\varphi\\
     0 &  -\sin\varphi& \cos\varphi
    \end{bmatrix}R,\label{eq2}
\end{equation}

where $ \varphi  $ is pitch angle.

The pitch angle is estimated by maximizing the cost $ c_{ \varphi } $ as follows,

\begin{equation}
\begin{aligned}
    c_{\varphi} = = \left \|\overline{a_1b_1} + \overline{a_2b_2} +\overline{a_3b_3} + \overline{a_4b_4}   \right \| ,\\ 
where \  \overline{a_ib_i} = \overrightarrow{a_ib_i} / \left \| \overrightarrow{a_ib_i} \right \| (i=1,2,3,4).\label{eq3}
\end{aligned}
\end{equation}

 As Figure 3 shown, point $ a_i $  and $ b_i $(i=1,2,3,4) correspond to the end points of the extracted markings in the surround-view image. $a_1$,$a_2$,$a_3$,$a_4$ are the intersection point with line $ y=0 $. $ \overline{a_{i}b_{i}} $ (i=1,2,3,4) is the unitized direction vector of $ \overrightarrow{a_{i}b_{i}} $, i.e., the direction vector of the lane marking. In this sub-step, we can approach the unique extreme value of  $ c_{\varphi} $ by iterating calculating rotation matrix with equation (2).

In the second sub-step, the iterative strategy is updating rotation matrix with, 

\begin{equation}
   R_{2}^{'} =R_{\omega  } R=\begin{bmatrix}
     \cos\omega & 0 & -\sin\omega\\
     0 & 1  & 0  \\
     \sin\omega &  0  & \cos\omega  
    \end{bmatrix}R,\label{eq4}
\end{equation}

Where $ \omega $ is yaw angle. In most cases, the width of the right lane line is equal to the left lane line. As a result, the yaw angle is estimated by minimizing the cost  $ c_{\omega} $ by,

\begin{equation}
    c_{\omega} =  \left |  \left |  a_{1}a_{2}  \right |  - \left |  a_{3}a_{4}  \right |   \right |  .\label{eq5}
\end{equation}

Where $ \left |  a_{1}a_{2}  \right | $ corresponds to length of $a_1a_2$ and $\left |  a_{3}a_{4}  \right |$  corresponds to length of $a_3a_4$. Similarly, we can approach the only extreme value of  $c_{\omega}$ by iterating the rotation matrix with equation (4).

In the third sub-step, the iterative strategy can be expressed as, 

\begin{equation}
   R_{3}^{'} =RR_{\theta   } =R\begin{bmatrix}
     \cos\theta    & -\sin\theta & 0   \\
      \sin\theta & \cos\theta  & 0  \\
      0 &  0  &  1  
    \end{bmatrix},\label{eq6}
\end{equation}

where $\theta$ is roll angle. Since the direction of the vehicle parallels to the lane marking, the projection line of the lane marking also parallels to the axis of the projection plane. The roll angle is estimated by minimizing the cost, $c_{\theta}$ as,

\begin{equation}
    c_{\theta} =  \textstyle \sum_{ i=1 }^{ n} \left | \cos\left \langle \widehat{a_{i}b_{i}} ,\overrightarrow{e_{1}} \right \rangle   \right |  ,\label{eq7}
\end{equation}

where  $ \overrightarrow{e_{1}} $ is the vector equals [1, 0] in the surround-view plane and n equals 4. Similarly, we can approach the only extreme value of  $c_\theta$ by iterating the rotation matrix with equation (6).

Overall, the motivations for each of the "cost functions" is not well described Eventually, it becomes clear that the first makes all the lines "point the same direction", the second normalizes the line widths, and the third makes the lines point "up" in the image. The total algorithm is showed in Algorithm 1.

\begin{algorithm}
\caption{Single Camera Pose Correction}\label{algo1}
\begin{algorithmic}[1]
\Require $ P, R, K_1, K_2 $
\Ensure $R$
\State Function
\State $ c_{angle}\gets F(R,P,K_1,K_2) $ angle $ \in \left \{  \varphi, \omega, \theta  \right \}  $
\State
\State Initialize $ R,P,K_1,K_2 $
\While{$iter < iter\_max $}
    \While{$c_{\varphi} < c_{\varphi,last} $}
        \State $ R\gets R_{\varphi}R $
        \State $ c_{\varphi}\gets F(R,P,K_1,K_2) $
    \EndWhile
    \While{$c_{\omega} < c_{\omega,last} $}
        \State $ R\gets R_{\omega}R $
        \State $ c_{\omega}\gets F(R,P,K_1,K_2) $
    \EndWhile
    \While{$c_{\theta} < c_{\theta,last} $}
        \State $ R\gets RR_{\theta} $
        \State $ c_{\theta}\gets F(R,P,K_1,K_2) $
    \EndWhile
    \State $ iter \gets iter + 1$
\EndWhile
\State return $ R $
\end{algorithmic}
\end{algorithm}

In the sub-step, phi, omega, and theta will be updated by determining the gradient direction of cost function. Ignoring the error of line detection, the first derivative in $c_\varphi$  of $\varphi$ is monotonic, so as to the derivative in $c_\omega$  of $\omega$, derivative in $c_\theta$  of $\theta$. Each sub-step iteratively approaches the extreme value of the cost functions and camera pose are able to be gradually estimated. 
 
Based on these cost functions, camera pose can be roughly estimated. This method can achieve an accurate camera pose estimation under precise lane marking detection. Considering the complex traffic scenario in reality, the line detection needs to be further refined in the subsequent step.

\subsection{Lane marking refinement}\label{subsec3}

The accuracy of lane marking detection is crucial for calibration. For the left and right cameras, since lane marking is quite close to the vehicle, the detection method mentioned in section.3.1 is sufficient to meet the accuracy requirements. However,  environmental conditions, such as extreme light and fragmentary lane marking, the above detection method yields unsatisfactory result for the front and rear cameras. 

As shown in Figure 4(b), the method is unstable in edge detection when lane marking is far away from the camera. This problem can be solved under the projection plane. After the rough calibration based on the lane line cost function, the preliminary estimated pose is utilized to generate a projection image based on homography. As illustrated in Figure 4(c), in the projection image, lane marking is accurately estimated by line detection, such as Hough line detection. The originally selected lane line is updated by searching for the nearest detected line. After correcting lane marking, the front and rear cameras are re-calibrated base on the lane marking cost function.

\begin{figure}[h]%
\centering
\includegraphics[width=1.0\textwidth]{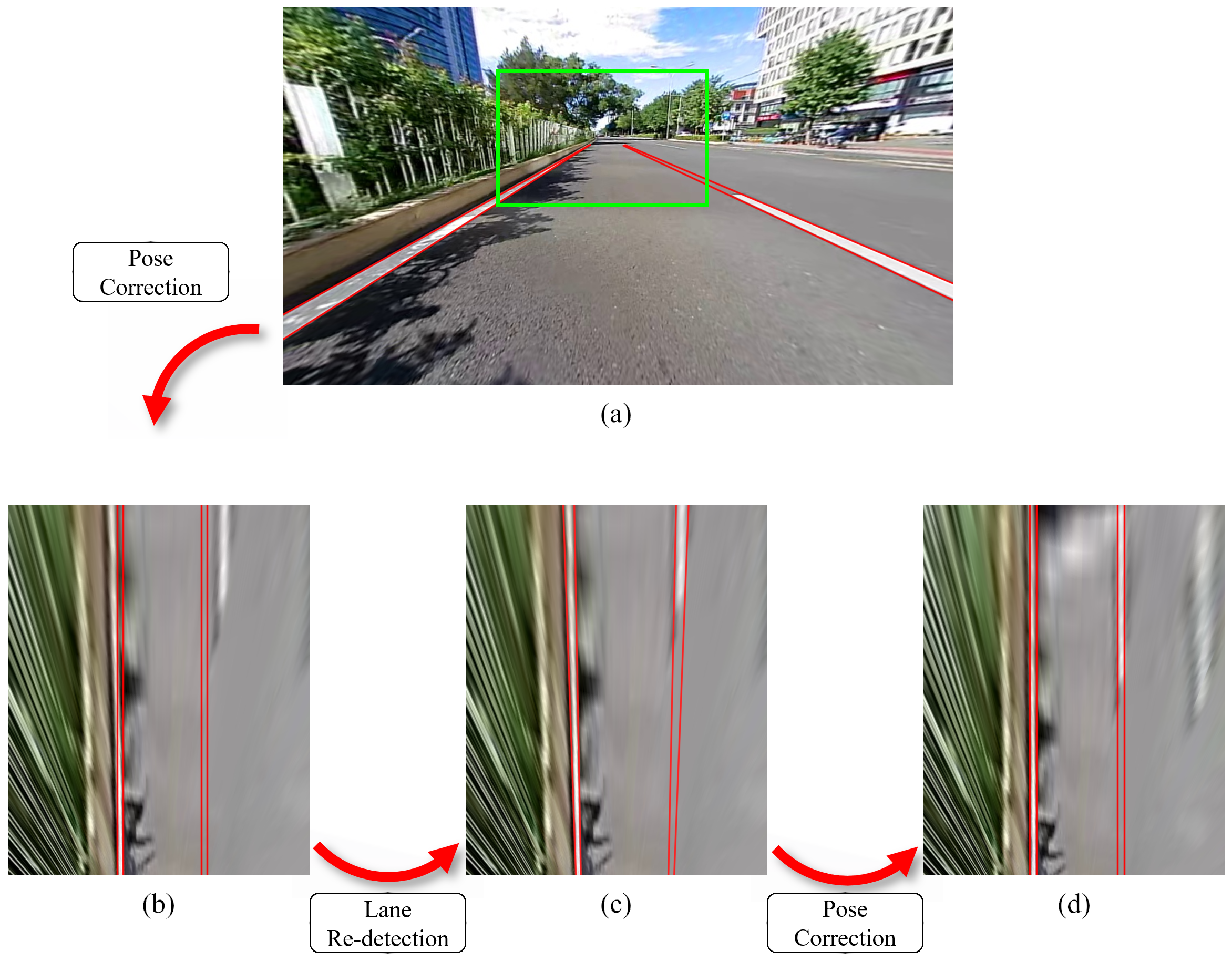}
\caption{(a) The straight lane line found in the undistorted image. (b) A rough projection after a rough calibration of the front camera. (c) Lane marking correction in the rough projection images.(d) After pose re-correction.}\label{fig3}
\end{figure}

\subsection{ Co-refinement of front and rear camera}\label{subsec3}

As shown in Figure 5, after the pose of front and rear cameras are respectively estimated, the x coordinate of the front and rear cameras in the world coordinate system can be optimized.

\begin{figure}[h]%
\centering
\includegraphics[width=0.4\textwidth]{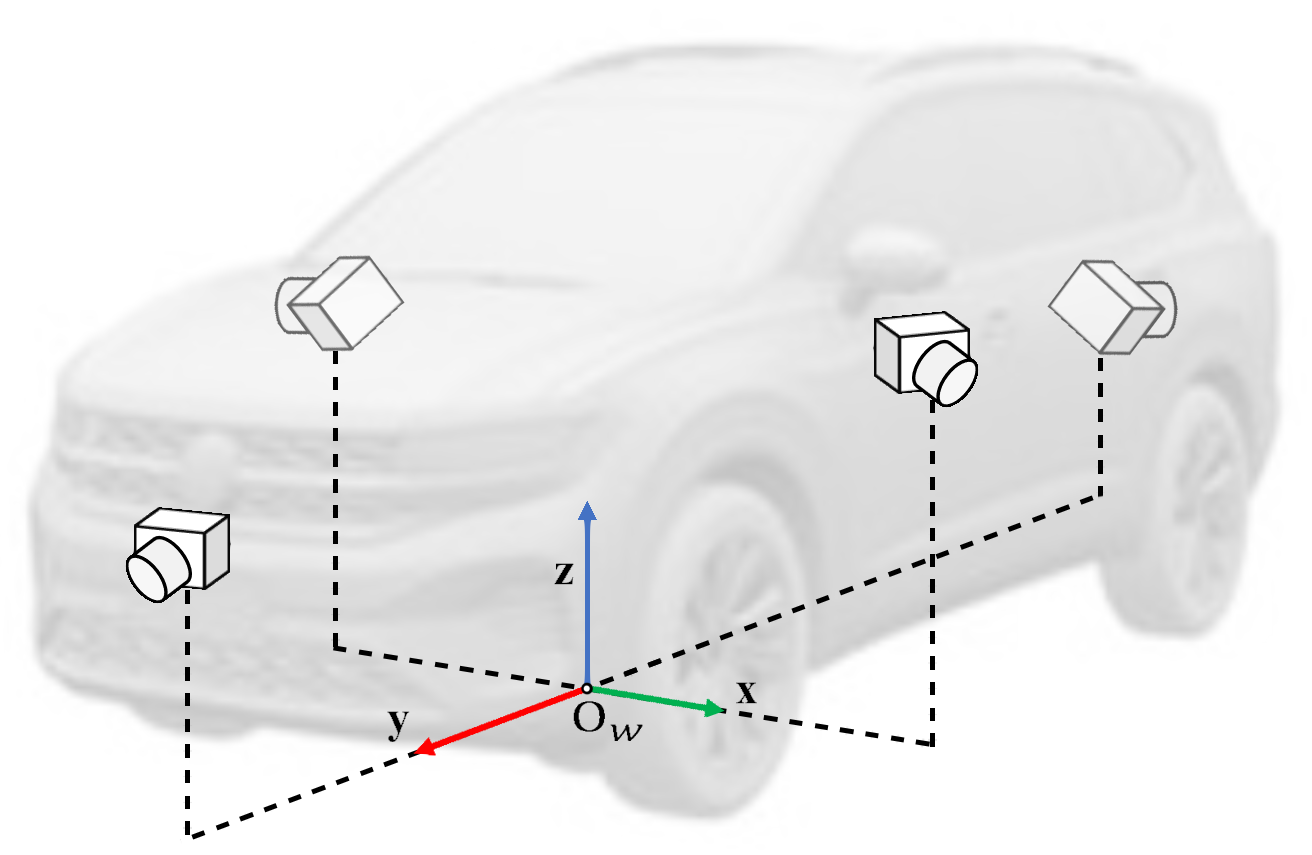}
\caption{world coordinate system of SVS. The y-axis direction is the direction of vehicle. The original point is the cross center of four cameras on the ground plane.}\label{fig5}
\end{figure}

As shown in Figure 6, in order to robustly generate the surround-view images, camera position error and pose error need to be taken into account. Therefore, the front and rear cameras will be simultaneously refined by minimizing the cost, $ c_{fb} $, as  

\begin{equation}
    c_{fb} =  c_{\omega,front} + c_{\omega,behind} +  {\textstyle \sum_{i=1}^{n}} \left | x_{fi}-x_{bi} \right |   ,\label{eq8}
\end{equation}

where $ x_{fi} $ , $ x_{bi} $ is x coordinate of $ a_{fi}$ and $ a_{bi} $ (i=1,2,3,4). The cost function $c_{fb}$ reflected the relative aligned error of front and rear camera. Therefore, we only optimizing the x coordinate of rear camera in the world coordinate system. 

\begin{figure}[h]%
\centering
\includegraphics[width=0.4\textwidth]{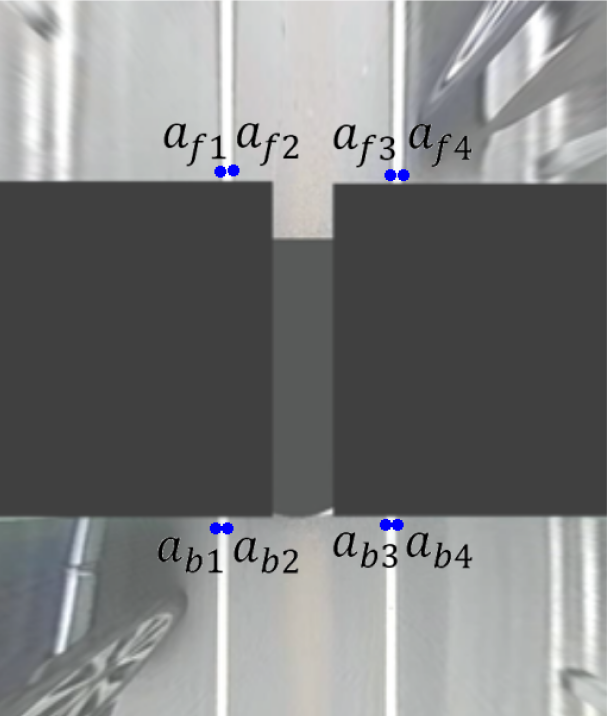}
\caption{Explanations of $ c_{fb} $  cost functions. The image of front and rear camera will be projected to the surround-view plane. The positions will be approached by aligning lane lines}\label{fig6}
\end{figure}

\subsection{ Calibrate left and right cameras }\label{subsec3}

The calibration of the left and right cameras is similar to that of the front and rear cameras. 

\begin{figure}[h]%
\centering
\includegraphics[width=1.0\textwidth]{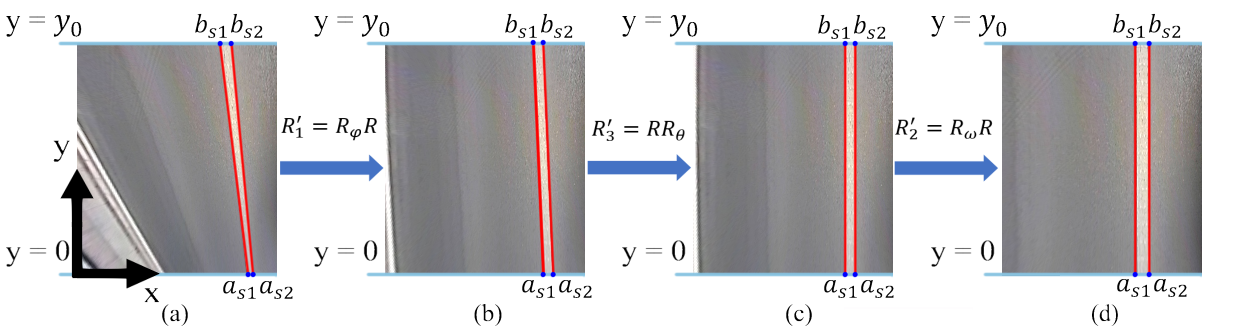}
\caption{Algorithm of pose estimation for front and rear camera. (a) The original surround-view.  (b). Sub-step of approaching yaw angle. (c) Sub-step of approaching roll angle . (d) Sub-step of approaching pitch angle.}\label{fig7}
\end{figure}

As Figure 7 shown, the method firstly estimates the yaw (Figure 7, (a) to (b)) and roll angles (Figure 7, (b) to (c)) of the side camera. The yaw and roll angles of the side camera are similarly estimated by the  method mentioned in Section 3.1. 

As Figure 7 shown, point $a_{s1}$, $a_{s2}$, $b_{s1}$, $b_{s2}$ respectively correspond to the end points of the extracted markings in the surround-view image, i.e., the intersection point with line $y=0$ and line $y=y_0$.

The yaw angle is estimated by minimizing the cost  $c_{\omega,side}$ by,

\begin{equation}
    c_{\omega,side} =  \left |  \left |  a_{s1}a_{s2}  \right |  - \left |  b_{s1}b_{s2}  \right |   \right |  ,\label{eq9}
\end{equation}

Where $\left |  a_{s1}a_{s2}  \right | $ corresponds to length of $a_{s1} a_{s2}$ and $\left |  b_{s1}b_{s2}  \right |$ corresponds to length of $b_{s1}b_{s2}$. Similarly, we can approach the only extreme value of  $c_{\omega,side} $by iterating the rotation matrix with equation (4).

The roll angle is estimated by maximizing  the cost, $c_{\theta,side}$ as,

\begin{equation}
    c_{\theta,side} =  \cos\left \langle \widehat{a_{s1}b_{s1}} ,\overrightarrow{e_{1}} \right \rangle + \cos\left \langle \widehat{a_{s2}b_{s2}} ,\overrightarrow{e_{1}} \right \rangle ,\label{eq7}
\end{equation}

Where $\widehat{a_{s1}b_{s1}}$, $\widehat{a_{s2}b_{s2}}$ respectively correspond to the unitized direction vector of $\overrightarrow{a_{s1}b_{s1}}$ and $\overrightarrow{a_{s2}b_{s2}}$, i.e., the direction vector of the lane marking. $\overrightarrow{e_{1}}$ is the vector equals [1, 0] in the surround-view plane and. Similarly, we can approach the only extreme value of  $c_{\theta,side}$ by iterating the rotation matrix with equation (6).

The optimization of pitch angle relies on the calibration results of the front and rear cameras. once the estimation of front and rear camera is completed, the width of lane marking can be simply calculated. The pitch angle is estimated by minimizing the cost  $c_{\varphi,side}$ by,

\begin{equation}
    c_{\varphi,side} =  \left |  \left |  a_{s1}a_{s2}  \right |  - d_{l}   \right |  ,\label{eq11}
\end{equation}

Where $\left |  a_{s1}a_{s2}  \right |$ corresponds to length of $a_{s1}a_{s2}$ and $d_l$ corresponds to the width of lane marking calculated in previous the section. Similarly, we can approach the only extreme value of  $c_(\varphi,side)$ by iterating the rotation matrix with equation (2).

However, due to the limited lane width accuracy of front and rear calibration, pitch estimation is unstable. This represents a limitation of calibration based on lane marking. As a result, a texture-based method is proposed to robustly estimate the pitch of the side camera.

\subsection{ Co-refinement of left and right camera }\label{subsec3}

In this step, the pipeline estimates pitch and position by contracting the pixel color error of RoI in the projection plane. Figure 8  provides an example of how adjacent cameras are processed by RoI texture. The texture alignment error considers  the color difference and edge gradient of pixels. Before calculating the pixel color error, the projected images of adjacent cameras need to be normalized on three color channels, which reduces the impact of adjacent cameras under different lighting conditions. 

\begin{figure}[h]%
\centering
\includegraphics[width=0.9\textwidth]{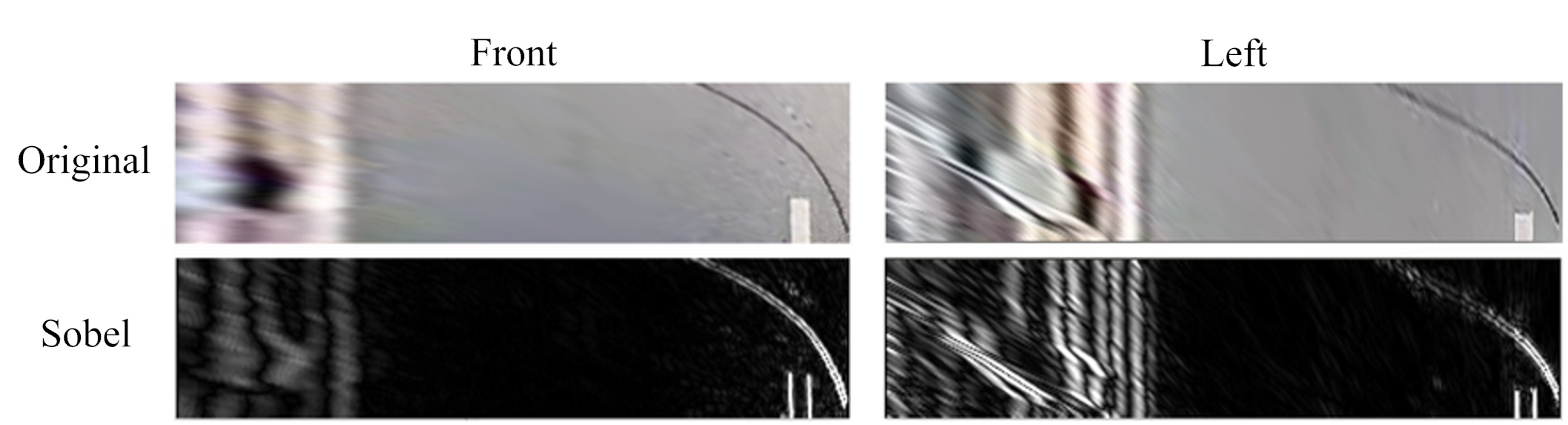}
\caption{Comparison of the adjacent camera projection. The pitch angle and position of bilateral camera can be estimated by traversal algorithm.}\label{fig8}
\end{figure}

Assuming that the boxed projection map of two adjacent cameras is $I_2$  and $I_1$. Normalization from $I_2$  to $I_1$ is,

\begin{equation}
    I_2^{'} = \frac{\sigma_1 }{\sigma_2}(I_2 - \mu_2) + \mu_1   ,\label{eq9}
\end{equation}

where $ \sigma_1 $, $ \mu_1 $ are standard deviation and mean value of $ I_1 $. $ \sigma_2 $, $ \mu_2 $ are standard deviation and mean value of $ I_2 $.

The pitch angle and position can be estimated by minimizing the cost, $ c_{text} $ as,

\begin{equation}
   c_{text}= \frac{1}{hw} {\textstyle \sum_{i=1}^{h}}  {\textstyle \sum_{j=1}^{w}} G_y(i,j)(I_1(i,j)-I_2^{'}(i,j))^{2}     ,\label{eq 10}
\end{equation}

where $h$ is the height of the RoI image. $w$ is the width of the image, $G_y$ is the longitudinal edge gradient of the pixel, $G_y$ is calculated by Sobel operator. Considering the characteristics of the fisheye lens, the clarity of projection images of two adjacent cameras is different. Therefore, $G_y$ is calculated by the camera whose RoI is closer to its principal point. For our dataset, we select left and right camera to calculate $G_y$. Due to the rough estimation of the previous step, the traversal range has been reduced. The pitch angle and position can be approached by traversal algorithm and the time cost is acceptable.

\subsection{ Parameter summarizing in sequence}\label{subsec3}

Sections 3.1-3.6 introduce the calibration at a certain moment. To apply the calibration to a long image sequence, the camera parameters are repeatedly estimated whenever a sufficient number of lane markings are collected. Using this approach, multiple parameter sets are obtained. A parameter set includes 12 camera angles (three angles for each camera) and can be represented as a 12-dimensional vector. To select the most appropriate parameter set and prevent overfitting from  a specific place, this paper uses the mean parameter set $ a^{'} $ of multiple parameter sets($ a_1 $,$ a_2 $,…,$ a_N $) as,

\begin{equation}
   a^{'} = \min_{b\in \{ a_1,a_2,\dots ,a_N \} } {\textstyle \sum_{i=1}^{N}} k_id(a_i,b)     ,\label{eq11}
\end{equation}

where $ N $ is the number of parameter sets, and $ d(a_i,b) $ indicates the Euclidean distance between two parameter sets ($ a_i $ and $ b $). $ k_i $ is the confidence as,

\begin{equation}
   k_i = \frac{c_{i,text}}{ {\textstyle \sum_{j=1}^{N}}c_{j,text} } , i=1,2,3,\cdots,N      ,\label{eq8}
\end{equation}

where $ c_{(i,text)} $ is the texture cost function of a time stamp.

\section{Our Dataset}\label{sec4}

Our data set was collected by four fisheye cameras mounted on cars. The dataset consists of 40 groups of videos. Each group of videos includes four video streams, front, back, left, and right. The frame rate is 25 fps.

Our dataset is captured by multiple sets of fisheye cameras with different camera parameters, and the camera system is installed on several vehicles for collection. As a result,  the GT of the extrinsic parameters in the dataset are various, as illustrated in Figure 9.

\begin{figure}[h]%
\centering
\includegraphics[width=1.0\textwidth]{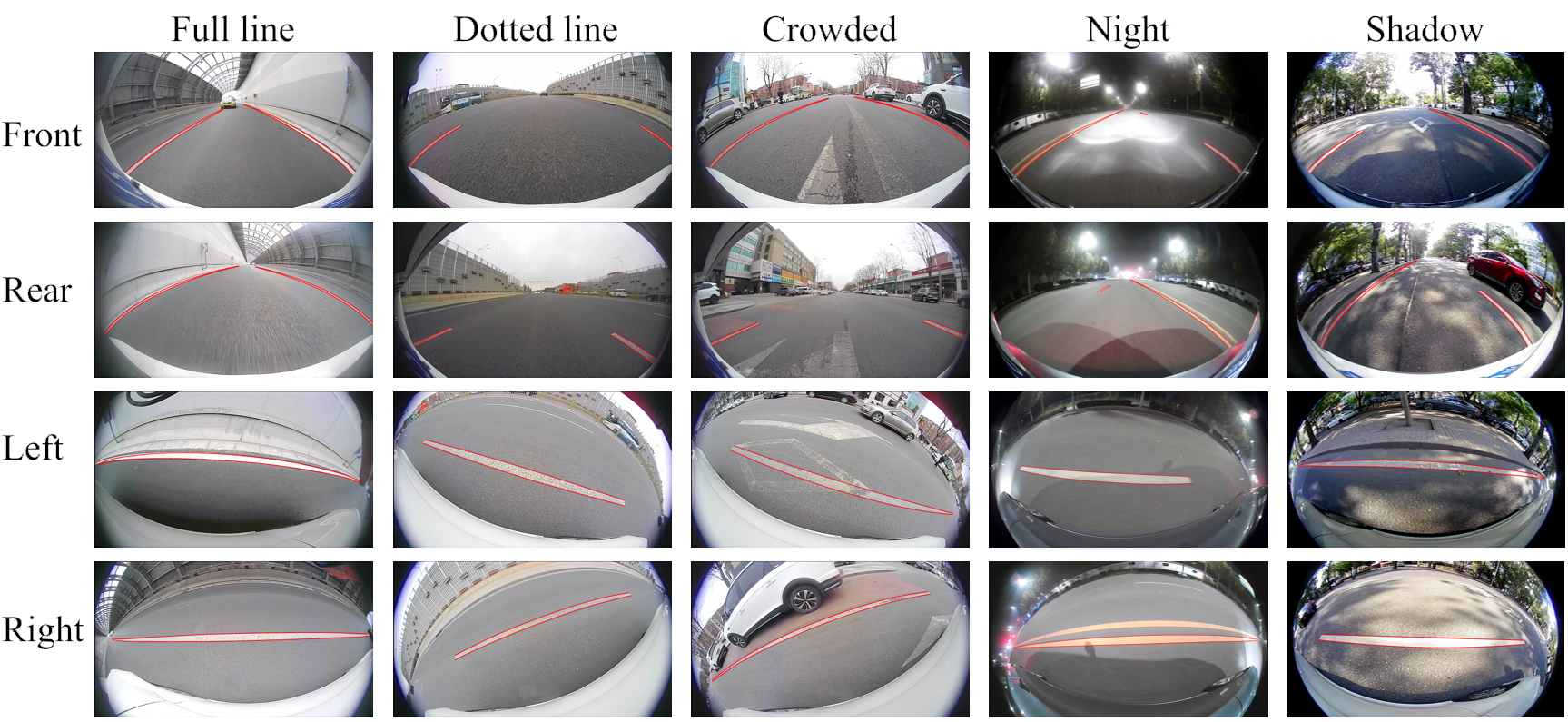}
\caption{The proposed video surround-view dataset contains different real traffic scenarios, and provides the lane annotations.}\label{fig9}
\end{figure}

The frames of each video share the same set of extrinsic parameters and GT, which are estimated using a chessboard.

The type of cameras is 2 million pixel high-definition panoramic waterproof lens from Shenzhen Zhongke Nanguang Technology Co., Ltd. The FOV of the camera is 190 degrees and the Image sensor type is IMX307 from Sony.

Lane lines, as well as the vehicle intrinsic and extrinsic parameters of each frame, are stored in the corresponding JSON file. Besides, all lane line annotations are manual. Each lane line is represented by two sets of edge points and is marked with the its  type, which includes single white solid line, single white dotted line, single yellow solid line, single yellow dotted line, double white solid line, double yellow solid line, double yellow dotted line, double white solid dotted line, double white solid line, white yellow solid line.

\section{Experiments}\label{sec5}

In real-world traffic scenario, when driving in the dotted lane, four cameras cannot capture the valid lane at the same time. Our dataset includes the solution. We reserve the lane lines that appeared in a short period. This is because driving direction can be assumed to be straight for a short period when diving to a straight lane.

We tested the pipeline with our dataset. We select clips from various environments as input. Partial results are shown in Figure 10. As can be seen from the results, our method has stable performance in various traffic scenarios.

\begin{figure}[h]%
\centering
\includegraphics[width=1.0\textwidth]{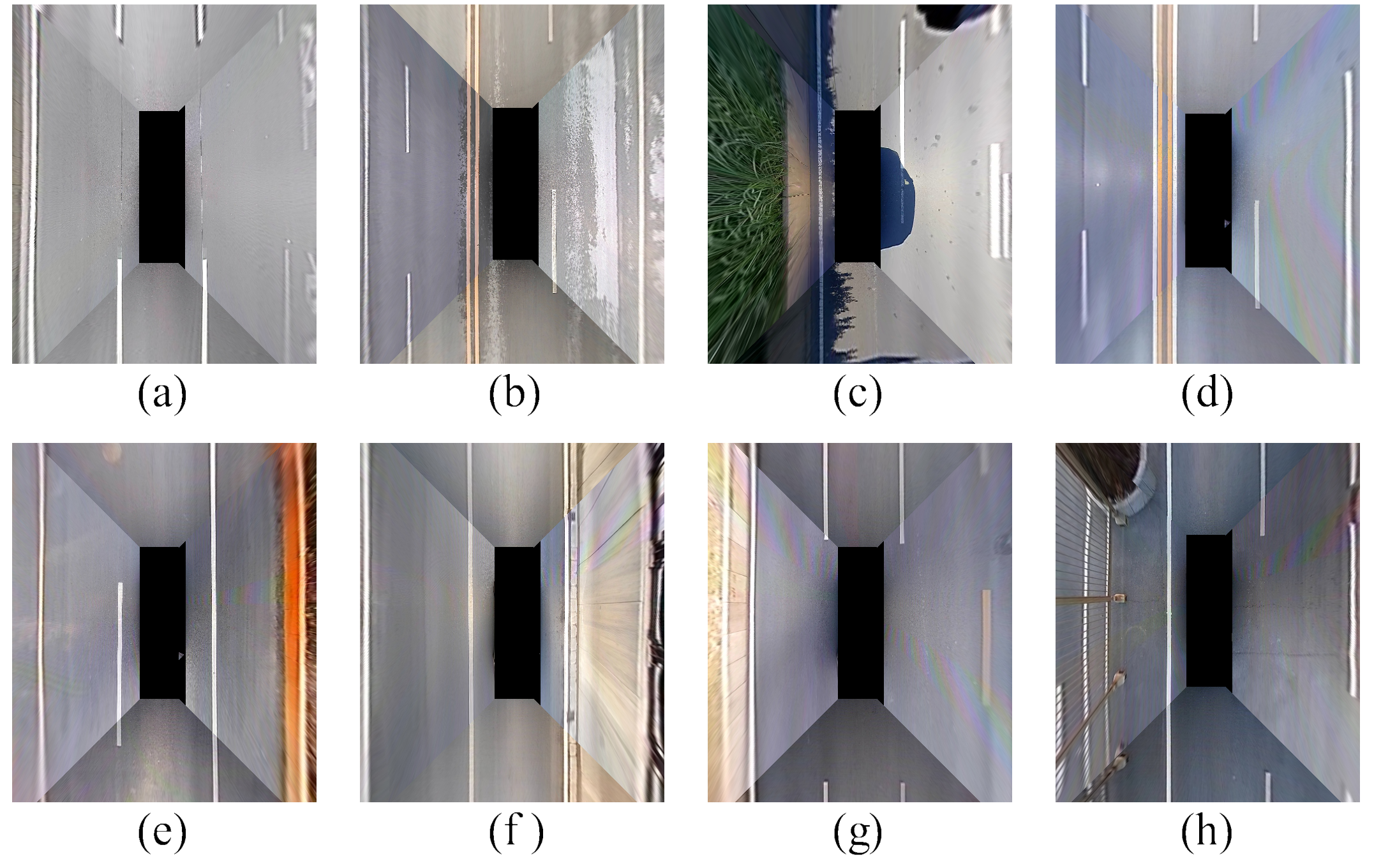}
\caption{Examples of representative traffic scenarios. Note that there is a wall or a footstep on the side of the vehicle in (c),  (e), (f) and (h). The 3D point that is not on the ground plane will be transformed incorrectly. }\label{fig10}
\end{figure}

In order to verify the effectiveness of the proposed method, we reproduce the scheme of Choi et al. \cite{bib24}, OCPO\cite{bib25} and WESNet\cite{bib3} for comparison. Each method initializes the position parameter with 5-centimeter error from GT in each world axis. Table 1 shows the result. the average camera angle error equals the average of three Euler angle. The average position error equals Euclidean distance between estimation and ground truth in the world coordinate system.

we provide detailed evaluation results of the calibration methods for each type of scenario. Table 1 shows the average angle error and average position error. The angle unit and length unit are degree and meter. 

The best results are highlighted in bold. From Table 1, it can be seen that compared with all counterparts, our method shows an overwhelming performance on all data groups. Choi’s method utilized lane marking and only estimated the camera angles. For the left and right camera, Choi’s method performs unstably due to the error of camera height. In contrast, our method estimates height and pitch together by constraining the width of projected lane marking and aligning the projected texture. Specifically, OCPO only takes the photometric error as the guidance to correct camera poses. However, in low-texture environment, the photometric error is mainly affected by noise rather than inaccurate poses of cameras. In this case, OCPO can yield extrinsic parameters with low photometric errors but their accuracy cannot be guaranteed.WESNet follows a weakly supervised learning framework. In the fine-tuning stage, WESNet utilizes photometric error to train the network, which will result in similar errors to OCPO.  In conclusion, the excellent accuracy of extrinsic parameter estimation and the generalization capability of our method has been nicely demonstrated.

\begin{table}[h]\tiny
\caption{Average Angle Error (degree) and Average Position Error (meter) for different methods}\label{tab1}

\begin{tabular*}{\textwidth}{@{\extracolsep\fill}llcccccccc}
\toprule

\multirow{2}{*}{Group} & \multirow{2}{*}{Method} 
& \multicolumn{2}{@{}c@{}}{Front camera} 
& \multicolumn{2}{@{}c@{}}{Rear camera} 
& \multicolumn{2}{@{}c@{}}{Left camera} 
& \multicolumn{2}{@{}c@{}}{Right camera} 
\\\cmidrule{3-4}\cmidrule{5-6}\cmidrule{7-8}\cmidrule{9-10}%

& & Angle & Position & Angle & Position & Angle & Position & Angle & Position \\

\midrule
\multirow{4}{*} {Full line} 
 & Choi $et$ $al$.\cite{bib24}  & 0.261 & none & 0.279 & none & 1.542 & none & 2.964 & none\\
 & OCPO\cite{bib25}  & 0.631  & 0.019  & 0.504  & 0.021  & 0.702 & 0.049 & 0.737 & 0.037\\
 & WESNet\cite{bib3}  & 0.369  & 0.030  & 0.265  & 0.019  & 0.424 & 0.029 & 0.465 & 0.021\\
 & Ours\footnotemark  & {\textbf {0.136}}  & {\textbf {0.003}}  & {\textbf {0.150}}  & {\textbf {0.006}}  & {\textbf {0.213}} & {\textbf {0.009}} & {\textbf {0.253}} & {\textbf {0.009}}\\
\midrule
\multirow{4}{*} {Dotted line} 
 & Choi $et$ $al$.\cite{bib24}  & 0.337 & none & 0.330 & none & 2.974 & none & 3.513 & none\\
 & OCPO\cite{bib25}  & 0.834  & 0.029  & 0.688  & 0.050  & 1.038 & 0.088 & 1.229 & 0.051\\
 & WESNet\cite{bib3}  & 0.501  & 0.032  & 0.549  & 0.038  & 0.870 & 0.040 & 0.989 & 0.024\\
 & Ours\footnotemark  & {\textbf {0.153}}  & {\textbf {0.004}}  & {\textbf {0.233}}  & {\textbf {0.005}}  & {\textbf {0.206}} & {\textbf {0.010}} & {\textbf {0.349}} & {\textbf {0.011}}\\
\midrule
\multirow{4}{*} {Crowded} 
 & Choi $et$ $al$.\cite{bib24}  & 0.289 & none & 0.291 & none & 3.321 & none & 3.950 & none\\
 & OCPO\cite{bib25}  & 0.977  & 0.035  & 0.840  & 0.074  & 1.481 & 0.122 & 1.916 & 0.088\\
 & WESNet\cite{bib3}  & 0.884  & 0.036  & 0.958  & 0.057  & 1.833 & 0.099 & 1.259 & 0.087\\
 & Ours\footnotemark  & {\textbf {0.326}}  & {\textbf {0.012}}  & {\textbf {0.398}}  & {\textbf {0.011}}  & {\textbf {0.608}} & {\textbf {0.018}} & {\textbf {0.631}} & {\textbf {0.017}}\\
\midrule
\multirow{4}{*} {Night} 
 & Choi $et$ $al$.\cite{bib24}  & 1.856 & none & 2.769 & none & 3.229 & none & 4.125 & none\\
 & OCPO\cite{bib25}  & 1.109  & 0.057  & 1.889  & 0.110  & 2.538 & 0.139 & 1.638 & 0.117\\
 & WESNet\cite{bib3}  & 0.835 & 0.087  & 1.594  & 0.084  & 2.217 & 0.108 & 1.526 & 0.062\\
 & Ours\footnotemark  & {\textbf {0.781}}  & {\textbf {0.062}}  & {\textbf {0.960}}  & {\textbf {0.022}}  & {\textbf {1.013}} & {\textbf {0.080}} & {\textbf {1.097}} & {\textbf {0.055}}\\
\midrule
\multirow{4}{*} {Shadow} 
 & Choi $et$ $al$.\cite{bib24}  & 0.308 & none & 0.323 & none & 3.591 & none & 3.911 & none\\
 & OCPO\cite{bib25}  & 0.529  & 0.031  & 0.497  & 0.063  & 1.517 & 0.109 & 1.270 & 0.077\\
 & WESNet\cite{bib3}  & 0.448  & 0.033  & 0.341  & 0.036  & 1.381 & 0.082 & 1.053 & 0.043\\
 & Ours\footnotemark  & {\textbf {0.257}}  & {\textbf {0.011}}  & {\textbf {0.146}}  & {\textbf {0.009}}  & {\textbf {0.534}} & {\textbf {0.027}} & {\textbf {0.398}} & {\textbf {0.021}}\\
\midrule
\multirow{4}{*} {Average} 
 & Choi $et$ $al$.\cite{bib24}  & 0.371 & none & 0.424 & none & 2.710 & none & 3.484 & none\\
 & OCPO\cite{bib25}  & 0.763  & 0.025  & 0.687  & 0.049  & 1.244 & 0.086 & 1.217 & 0.060\\
 & WESNet\cite{bib3}  & 0.486 & 0.035  & 0.513  & 0.037  & 0.965 & 0.053 & 0.887 & 0.030\\
 & Ours\footnotemark  & {\textbf {0.228}}  & {\textbf {0.009}}  & {\textbf {0.262}}  & {\textbf {0.008}}  & {\textbf {0.376}} & {\textbf {0.017}} & {\textbf {0.398}} & {\textbf {0.015}}\\
\botrule
\end{tabular*}
\end{table}

\begin{table}[h]
\caption{Result of ablation experiment}\label{tab1}
\begin{tabular*}{\textwidth}{@{\extracolsep\fill}lcccccccc}
\toprule%
& \multicolumn{2}{@{}c@{}}{Front camera} & \multicolumn{2}{@{}c@{}}{Rear camera} & \multicolumn{2}{@{}c@{}}{Left camera} & \multicolumn{2}{@{}c@{}}{Right camera} \\\cmidrule{2-3}\cmidrule{4-5}\cmidrule{6-7}\cmidrule{8-9}%
Method & \begin{tabular}[c]{@{}c@{}}Angle\\ (degree)\end{tabular} & \begin{tabular}[c]{@{}c@{}}Position\\ (meter)\end{tabular} & Angle & Position & Angle & Position & Angle & Position \\
\midrule
Ours\footnotemark[1]  & 0.367  & 0.023  & 0.485  & 0.028  & 0.648 & 0.034 & 0.674 & 0.029\\
Ours\footnotemark[2]  & 0.228  & 0.009  & 0.262  & 0.008  & 0.554 & 0.076 & 0.597 & 0.058\\
Ours  & 0.228  & 0.009  & 0.262  & 0.008  & 0.376 & 0.017 & 0.398 & 0.015\\
\botrule
\end{tabular*}
\footnotetext[1]{without lane marking re-detection.}
\footnotetext[2]{without texture procedure.}
\end{table}

In Table 2, we conducted the ablation experiment to prove the effectiveness of each component we propose. The lane marking re-detection improves extrinsic estimation by enhancing lane marking detection accuracy of each camera. The texture procedure also contributes to calibration. 

\begin{figure}[h]%
\centering
\includegraphics[width=0.9\textwidth]{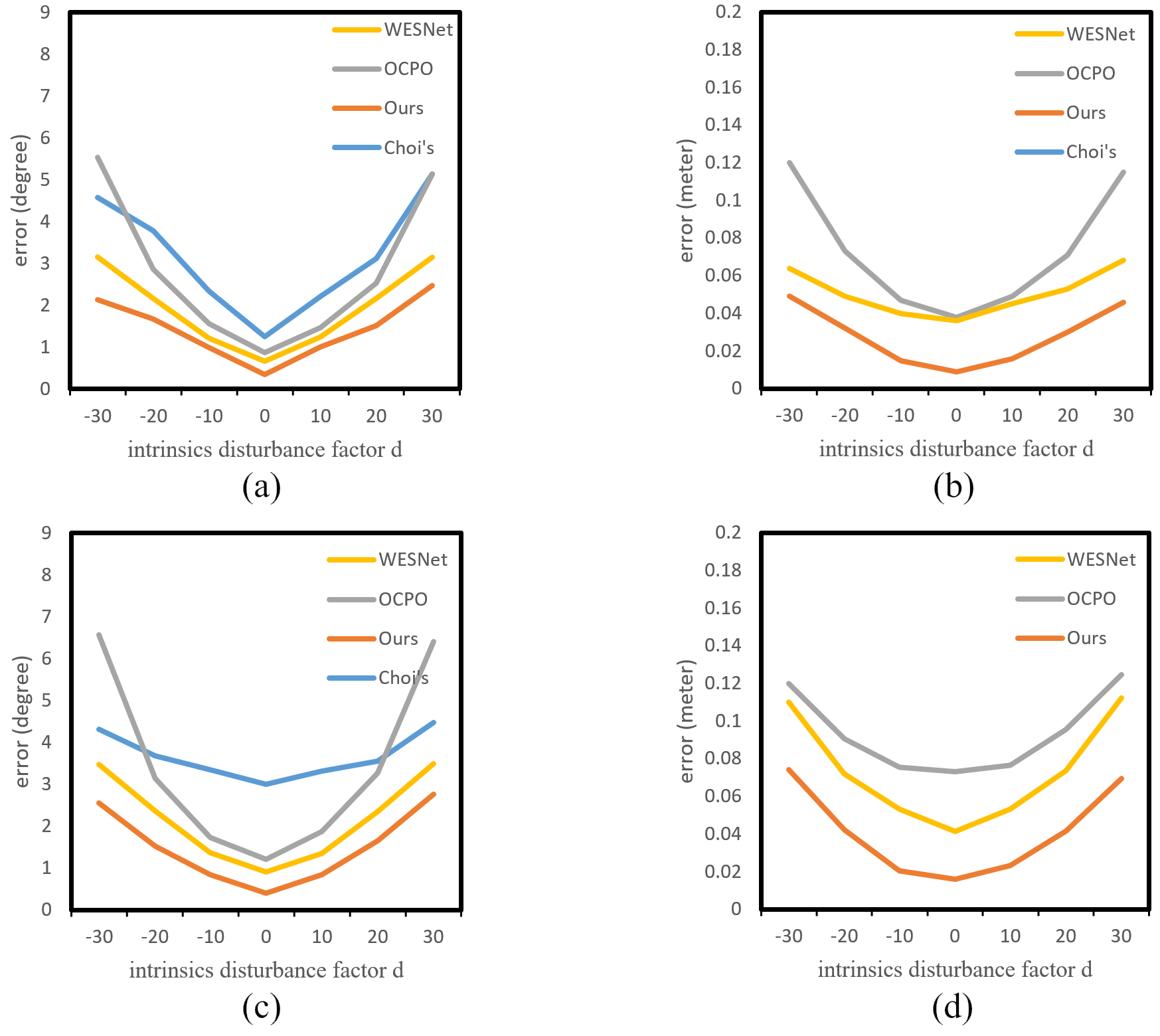}
\caption{angle error and position error under different disturbance factor d’s settings. (a) and (b) separately shows average angle error and position error of fonrt and rear camera. (c) and (d) separately shows the error of left and right camera.}\label{fig8}
\end{figure}

\bmhead{Robustness to intrinsic disturbance}In order to evaluate the robustness of our method of the accuracy of intrinsic parameters, we empirically alter the intrinsic parameters of each camera. The disturbance can be represented as an intrinsic disturbance factor $ d $ . We added it to the focal length of the camera. The disturbed intrinsic matrix $ K_{C_i}^d $ of camera $ C_i $ can be expressed as

\begin{equation}
   K_{C_i}^d = \begin{bmatrix}
  f_x+d& 0 & c_x \\
  0& f_y+d & c_y\\
  0& 0 &1
\end{bmatrix}  .\label{eq13}
\end{equation}

We test the mentioned method and our scheme with different $d$’s setting and recorded the result. The result is shown in Figure 11. It illustrates that our method is able to calculate a more accurate result as long as the intrinsic disturbance is higher. The reason is that the intrinsic parameter are disturbed, the pixel is farther from principal point and yields higher distortion error. Our scheme utilizes information about lane as a strong constraint. Lane lines mostly  distributed in the center of fisheye image while the RoI distributed at the side of fisheye  image. Therefore, lane-base calibration is more robust than texture-base calibration. Besides, based on our experience, the camera’s focal length variation caused by the natural collisions or bumps  is generally less than 5 pixels. Therefore, it can be concluded that our method is robust to the variations in intrinsic parameters.

\bmhead{Execution Time}Table 3 shows the execution time for four main modules of the proposed method. These times were measured on an Intel Core i7-10875 CPU using only a single core that has 2.30GHz. Note that among four modules the proposed method, only the lane marking detection module is required to be processed in real time. The other three modules need to be processed only once after a sufficient number of lane markings are gathered. In Table 3, the lane marking detection module requires 29.74 ms to process four images acquired from four cameras of the AVM system, which means that this module can process more than 30 frames per second in real time. The other three modules require a total execution time of 315.54 ms. Since those modules need to be processed only once after gathering lane markings, their execution times do not hinder the proposed method from operating in real time.

\begin{table}[h]
\caption{Execution time}\label{tab2}
\begin{tabular}{cc}
\toprule
Module                 & Time(ms) \\
\midrule
Lane marking detection & 29.74    \\
Lane marking selection & 3.89     \\
Parameter estimation   & 311.63   \\
Parameter selection    & 0.02     \\
\botrule
\end{tabular}
\end{table}

\section{Conlusion}\label{sec6}

In this paper, a new dataset and practical method are proposed to calibrate the camera attitude using lane markings and textures. To facilitate the research on self-calibration for the surround-view system, we collected a new dataset with high-quality lane annotation across all the frames. The proposed method offers several advantages. First, it is suitable for the natural driving situation of vehicles, whether the vehicle is moving at low speed or high speed. Second, the method uses the corresponding lane lines from  adjacent camera images to relate each camera,  generating a seamless aerial view of vehicles. The method is evaluated using image sequences captured under various actual driving conditions, and the results demonstrate good real-time performance. 

The experiment uses a four-camera system. Therefore, this calibration method is also applicable to the six-camera system with six or more cameras. In the six-camera system, there will be two cameras at the front and two cameras at the rear of the vehicle when the vehicle is driving in the lane. Lane lines of two sides can be capture simultaneously.  By treating the camera that captures two lane lines are as the "front and rear views angles" in the calibration process, and the camera that captures one side lane line as the "left and right views angles", this method can be used for camera calibration and circular projection splicing.

\section{Data availability}
The datasets associated with the current study are available upon reasonable request from the corresponding author.

\section{Conflict of interest}
The authors certify that there are no actual or potential conflicts of interest in relation to this article.

\begin{appendices}




\end{appendices}


\bibliography{sn-bibliography}

\end{document}